%% file: main-arxiv.tex
\documentclass[runningheads,a4paper]{article}
\usepackage{arxiv}

\usepackage{authblk}
\usepackage{footnote} 
\usepackage{chemformula} 
\usepackage{cite}
\usepackage{amsmath,amssymb,amsfonts}
\usepackage{algorithmic}
\usepackage{graphicx}
\usepackage{textcomp}
\usepackage{xcolor}
\usepackage{todonotes}
\usepackage[fencedCode]{markdown}
\usepackage[
            binary-units=true,
			separate-uncertainty=true,
            ]{siunitx}
\usepackage{hyperref}
\usepackage{booktabs}

\usepackage{tabularx}
\newcolumntype{R}{>{\raggedleft\arraybackslash}X}%
\newcolumntype{L}{>{\raggedright\arraybackslash}X}%

\usepackage{floatrow}
\newfloatcommand{capbtabbox}{table}[][\FBwidth]

\begin{document}

\title{Implications of Noise in Resistive Memory on Deep Neural Networks for Image Classification}

\renewcommand\Authfont{\bfseries}
\setlength{\affilsep}{0em}
\newbox{\orcid}\sbox{\orcid}{\includegraphics[scale=0.06]{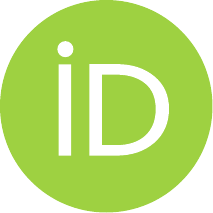}} 
\author[1]{%
    Yannick Emonds%
}
\author[2]{%
    Kai Xi\thanks{\texttt{xikai@ime.ac.cn}}%
}
\author[1]{%
    \href{https://orcid.org/0000-0000-0000-0000}{\usebox{\orcid}\hspace{1mm}Holger Fröning\thanks{\texttt{holger.froening@ziti.uni-heidelberg.de}}}%
}
\affil[1]{Computing Systems Group, Institute of Computer Engineering, Heidelberg University, Germany}
\affil[2]{Key Laboratory of Microelectronics Device \& Integrated Technology, Institute of Microelectronics of Chinese Academy of Sciences, China}

\maketitle

\begin{abstract}
Resistive memory is a promising alternative to SRAM, but is also an inherently unstable device that requires substantial effort to ensure correct read and write operations. To avoid the associated costs in terms of area, time and energy, the present work is concerned with exploring how much noise in memory operations can be tolerated by image classification tasks based on neural networks. We introduce a special noisy operator that mimics the noise in an exemplary resistive memory unit, explore the resilience of convolutional neural networks on the CIFAR-10 classification task, and discuss a couple of countermeasures to improve this resilience.
\end{abstract}

\input{01-introduction.tex}

\input{02-background.tex}
\input{03-rram_noise.tex}

\input{04-operators.tex}
\input{05-experiments.tex}
\input{06-resilience.tex}
\input{07-summary.tex}

\section*{Acknowledgments}

This work is funded by the Deutsche Forschungsgemeinschaft (DFG, German Research Foundation) under grant number 449797478.

\bibliographystyle{splncs04}

\bibliography{main}

\appendix
\newpage
\input{08-appendix.tex}

\end{document}

%% file: 01-introduction.tex
\section{Introduction}

Resistive memory (RRAM) is a promising memory technology to improve density and power consumption in comparison to SRAM. 
As resistive memory devices are inherently unstable, substantial engineering effort is required to minimize read and write uncertainty to acceptable levels, which also results in penalties in terms of performance and/or area.
Machine learning (ML) based on deep (artificial) neural networks is among the top methods for reasoning under uncertainty, and has advanced state-of-the-art performance in various prediction tasks including image, speech and signal processing. 
However, it comes at extreme costs in terms of memory and compute requirements. 
In contrast, it is often desirable to deploy ML methods on resource-constrained devices, including IoT, edge and mobile~\cite{Roth2020}. 
For such devices, RRAM can be a helpful technology to reduce the gap in between ML's requirements and the resources of the target device.

Excellent reviews of resistive memory and its use for different computations can be found in \cite{Yang2013} and \cite{7120149}, with technological insights summarized in Table~\ref{table:memory_technology}.
Various work proposes to use RRAM as building block for resistive-memory-based crossbars, in which a resistive memory cell performs a multiplication based on an input voltage (activation) and the programmable resistance of the memory cell (weight).
While such analog computations are highly promising due to an excellent energy efficiency,
there are various associated problems, most notably frequent A/D and D/A conversions.
Thus, this work is rather concerned with a digital use of resistive memory.
to focus on understanding the inherent noise tolerance in neural networks.
With regard to a more general treatment of noise and fault tolerance in machine learning, a notable body of work reports on implications of noisy computations for ML tasks~\cite{Zhou2020,item2021,electronics9030414,9654206} and proposes a set of remedy methods. 
In summary, one can observe that noise has been considered in some ML methods, but mainly for computations.

This work is thus concerned with noisy memory accesses as a result of the inherent instability of resistive memory, and will consider image classification tasks based on neural networks as application.
We will start with reviewing and discussing different sources of noise in resistive memory to facilitate an understanding of the most important noise sources. 
We continue by introducing a special PyTorch operator that mimics the noise found in such devices.
For an exploration of the implications on neural networks for image classification, we will consider a processor architecture that stores all parameters (i.e., all model weights and all activations) in RRAM.
Based on convolutional neural networks of different depth on the CIFAR-10 data set, we then explore accuracy-noise trade-offs to understand the implications in more detail.
Last, we shortly study the effect of selected techniques and methods to improve model resilience.

Considering a stuttering Moore’s law and diminished returns from CMOS technology scaling, new methods have to be found to maintain performance scaling. 
As tolerating noisy memory operations can improve RRAM’s performance while reducing its costs,
it is thus promising to understand the implications of such noisy memory operations for an improved performance scaling based on resistive memory.
In this regard, one key question that motivates the present work is how much noise in memory operations can be tolerated by plain ML architectures, and, as an early study, how much this noise tolerance is improved by resilience methods.

%% file: 02-background.tex
\section{Background}

\subsection{Resistive Memory}

RRAM, also known as memristor, typically refers to the resistive switching between different states controlled by the formation and rupture of conductive filaments inside of numerous metal oxides including zirconium oxide (ZrOx), nickel oxide (NiOx), hafnium oxide (HfOx), titanium oxide (TiOx), and tantalum oxide (TaOx). 
Various studies have demonstrated that the feature size of RRAM can be scaled down to sub-10nm \cite{Govoreanu2011a} and the device offers desirable electrical properties such as nanosecond switching speed \cite{Choi2016a}, long-term data retention, $10^{12}$ cycles endurance \cite{Kim2010a}, programming currents as low as nanoamperes \cite{Zhou2016a}, and large switch window, which describes the difference in between the low-resistive state (LRS) and high-resistive state (HRS), for data storage and logic calculation. 
Table~\ref{table:memory_technology} summarizes the most important characteristics of memory technologies.

Resistive memory is a promising memory technology, as it essentially brings the density of NAND FLASH while improving the corresponding read and write time by about four orders of magnitude.
Still, in comparison to other technologies like DRAM and SRAM, RRAM has two to four orders of magnitude higher energy costs per bit, which demands for careful optimizations when accessing RRAM.
However, the community expects much better future performance scaling for RRAM, as it is a nascent technology, in particular in comparison to established ones like SRAM and DRAM. 
Also, RRAM's manufacturing process is fully compatible with CMOS, allowing for tight integrations, such as in-memory computing.

\begin{savenotes}
\begin{table}
\small
\centering
\caption{Memory technology overview (adapted from~\cite{7120149} and~\cite{Yang2013})}
\begin{tabular}{ l | c | c | c | c | c | c  } 

							& SRAM				& DRAM					& NAND 				& PCM		& STT		& RRAM	\\
							& 					& 						& FLASH				& 			& RAM		& 		\\
\hline
Cell size $[F^2]$			& $140$				& $6-12$				& $1-4$				& $4-16$	& $20-60$	& $<4$		\\
Energy per bit $[pJ]$		& $5 \cdot 10^{-4}$	& $5 \cdot 10^{-3}$		& $2 \cdot 10^{-5}$	& $2-25$	& $0.1-2.5$	& $0.1-3$	\\
Standby power $[W]$			& N/A \footnote{Standby is not supported by SRAM}
												& refresh power			& $0$				& $0$		& $0$		& $0$		\\
Read time $[ns]$			& $0.1-0.3$			& $10$					& $10^5$			& $10–50$ 	& $10-35$	& $<10$		\\
Write time $[ns]$			& $0.1-0.3$			& $10$					& $10^5$			& $50-500$	& $10-90$	& $~10$		\\
Retention $[time]$			& $\infty$	\footnote{SRAM retention is unlimited as long as supply voltage is applied}		
												& $<< 1s$ 				& years				& years		& years		& years		\\
Endurance $[cycles]$		& $>10^{16}$		& $>10^{16}$			& $10^4$			& $10^9$	& $10^{15}$	& $10^{12}$	\\
Access granularity $[B]$ \footnote{\label{note1}Flexible access granularity refers to flexible at design time}	
							& flexible 			& $64$					& $4k$				& flexible 	& flexible 	& flexible		

\end{tabular}
\label{table:memory_technology}
\end{table}
\end{savenotes}

\subsection{Related Work}

There exists a plethora of work with regard to the use of RRAM in ML.
An In-Memory Data Parallel Processor based on RRAM with associated prototypical workflow is described by Fujiki et al. in \cite{Fujiki2018a}, making it an early but comprehensive work that demonstrates the potential of integrated RRAM for memory and computations. Note that neither noise nor deep neural networks are addressed in this work.
PRIME is a processing-in-memory architecture for neural networks based on resistive main memory~\cite{10.1109/ISCA.2016.13}.
While showing outstanding performance improvements compared to CPUs, it again does not address noise in resistive memory but assumes solely save memory operations.

ISAAC\cite{10.1145/3007787.3001139} is an example of an accelerator architecture in which the inherent dot products of neural networks are processed in crossbars based on resistive memory as connections.
This work, as well as various others more focused on such resistive-memory-based crossbars
are inherently analog forms of computations as weights are encoded in an analog fashion in the resistive connections.
As mentioned previously, the present work considers resistive memory in a digital fashion, thereby avoiding various problems associated with analog computations.
In particular, analog computing requires frequent conversions in between analog and digital domains to avoid infinite error propagation and accumulation.
The required circuitry in the form of A/D and D/A converters consumes substantial amounts of time, area and energy.

The impact of variability in resistance distributions has been explored in \cite{electronics9030414}, albeit only one single ImageNET image is considered and device variation is limited, too. 
Similar work has been done in \cite{9654206}, however here different convolutional operators are analyzed.
The work called "Noisy Machines" introduces a methodology based on knowledge distillation to highlight and combat the reduced learning capacity of ANNs when executed on noisy hardware~\cite{Zhou2020}.
Last, the inherent noise of analog computations has been analyzed in \cite{item2021} and a methodology for improved resilience by noisy training been proposed.

This work contrasts from previous work by focusing on the resilience of ANNs with regard to noise in memory accesses.
For that, we model the most important noise sources in an efficient manner using a PyTorch data structure called \emph{fliptensors}, which basically represents bit flips.
The low overhead of this noise model allows us in the following experiments to scale to deep model architectures and full data sets.

%% file: 03-rram_noise.tex
\section{Review of Noise Sources in Resistive Memory}
Non-idealities occur for different operations that are carried out on RRAM cells.
In this section, we briefly review and discuss the most important ones.

The most prominent source of noise during a read operation is the so-called \emph{Random Telegraph Noise} (RTN).
It is caused by temporary charge traps in the vicinity of the conductive filament of a RRAM cell \cite{Ambrogio2014b}.
These traps create a local electrical field thereby changing the conductivity of the filament.
The read resistance therefore varies over time.
As this effect highly depends on the size of the conductive filament, and thus on the resistance of a state, RTN affects the HRS much more than the LRS.
If the memory window between HRS and LRS is large enough, the states can still be well distinguished.

\begin{figure}
\centering
\includegraphics[width=1\textwidth]{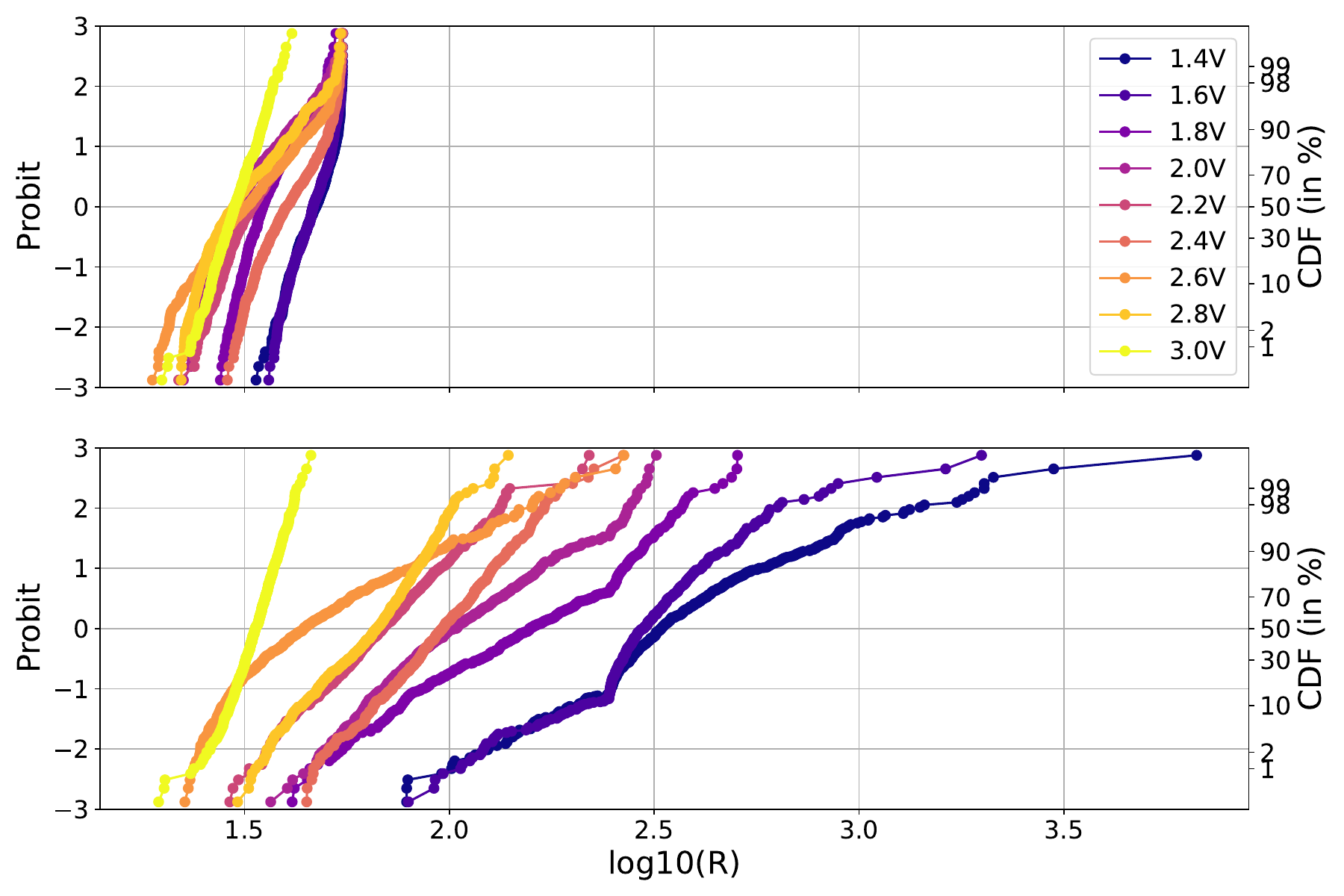}
\caption{Cycle-to-cycle variability for LRS (upper subfigure) and HRS (lower subfigure) measured on an HfOx test device for different forming voltages.}
\label{fig:c2c_var}
\end{figure}

Deviations of resistances can also arise when writing a RRAM cell.
The main effect causing them is called \emph{cycle-to-cycle variability} \cite{Ambrogio2014a, Fantini2013}.
Here, a cycle is the sequence of one set and one reset operation.
Stochastic variations within the conductive filament during the write operation lead to different resistance values for each cycle.
The result is a distribution of resistances similar to a lognormal distribution \cite{Karpov2017}.
Write noise is then the amount of overlap of the resistance distributions for HRS and LRS.
This is shown in figure \ref{fig:c2c_var} for an HfOx test device, reporting cumulative distribution function (CDF) and probit for different forming voltages.
HfOx material was chosen as it is among the most commonly used RRAM materials.
One can see which effect an increased forming voltage has on the CDF (and probit) of the resulting resistance.
Similar results, but for a different HfOx test device, have been reported by Fantini et al.~\cite{Fantini2013}.

The resistance window $\Delta RW$ as median-to-median distance between HRS and LRS distributions scales with current, however, larger $\Delta RW$ is desired as it increases robustness against the cycle-to-cycle variability.
For this data, only HRS and LRS distributions for the lowest two forming voltages distinguish completely.
All other distributions overlap to some extent, thus there is a probability that a LRS is interpreted as HRS and vice versa.

\subsection{Bit Flips} \label{subsec:bit_flips}
Lifting the level of abstraction from an analog to a digital perspective, it becomes clear that noise only appears as bit flips.
Only noise events that are large enough to cross the threshold of the ADC have to be taken into account.
Based on literature values for RTN, we conclude that this effect is too small to trigger a bit flip as the resistance states are usually of different orders of magnitude.
On the other hand, cycle-to-cycle variabilities exhibit an overlap of the resistance distribution that can get large enough to cause a bit flip.
Hence, in this work, we consider write noise due to cycle-to-cycle variabilities as the main source of noise.

To obtain the probabilities of the bit flips, it is necessary to choose a threshold resistance, from which one can decide if a cell is in HRS or LRS.
Since none of these states is preferable over the other, we set the threshold such that the probabilities for a flip from LRS to HRS and vice versa are equal.
This can be realized by requiring

\begin{equation}
1 - \mathrm{CDF}_\mathrm{LRS}\!\left(R_\mathrm{thresh}\right) \overset{!}{=} \mathrm{CDF}_\mathrm{HRS}\!\left(R_\mathrm{thresh}\right) =: p_\mathrm{bf} \, ,
\end{equation}
for the threshold resistance $R_\mathrm{thresh}$ and bit flip probability $p_\mathrm{bf}$.

\subsection{Midpoint Noise Level}
In order to establish a common metric that allows to compare the results for different configurations, we follow the notion of \emph{midpoint noise level} as proposed in \cite{Borras2022}.
The accuracy curve for a sweep over bit flip noise probabilities is typically shaped like a logistic curve.
Hence, a fit to a logistic function seems natural:
\begin{equation}
F\!\left(x; \mu, \sigma, \delta a, a_\mathrm{min}\right) = \frac{2 \cdot \delta a}{1 + e^{\left(x-\mu\right)/\sigma}} + a_\mathrm{min} \, .
\label{eq:logistic}
\end{equation}
Here, $\mu$ denotes the midpoint of the curve, hereafter the midpoint noise level.
Furthermore, $\sigma$ governs the slope of the curve, $a_\mathrm{min}$ is the asymptotic minimum, which for classification tasks with $K$ classes equals to $1/K$, and $a_\mathrm{max}$ is peak accuracy, usually achieved with the lowest amount of noise.
Finally, $\delta a = \left(a_\mathrm{max} - a_\mathrm{min}\right) / 2$ to compensate for the addition of $a_\mathrm{min}$ in equation \eqref{eq:logistic}.
Then, $\mu$ can be interpreted as a measure of model resilience, with higher values representing more resilience against noise.

%% file: 04-operators.tex
\section{Simulating Noise during Training and Inference}
In this section we outline a procedure to simulate bit flip noise within the PyTorch framework.
This is realized by implementing a new type of network layer whose task is to make access to the specified tensors noisy.

\begin{figure}
\centering
\includegraphics[width=.5\textwidth]{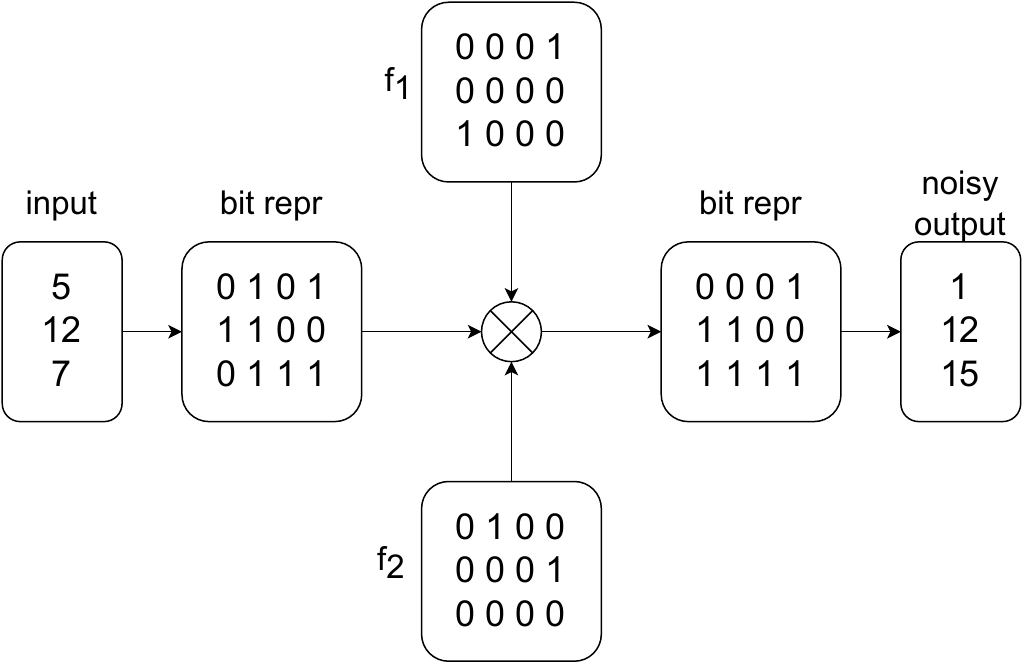}
\caption{Exemplary application of the fliptensors $f_1$ and $f_2$ to a vector with three 4-bit integer values.}
\label{fig:fliptensor}
\end{figure}

A bit flip can occur in one of two directions: from 0 to 1 or from 1 to 0.
In principle, the probabilities for these noise events can be different.
Let $p_1$ be the probability for a bit to flip from 0 to 1, and $p_2$ for the opposite.
Given the input tensor $\hat{x}$  with values of some data type, the bit flip noise is applied to a tensor $x$ containing the bit representation of the values of $\hat{x}$.
The naive approach is to iterate over the entire tensor, check the bit and flip it according to the respective probability $p_1$ or $p_2$.
However, alternatively, two tensors $f_1$ and $f_2$ of the exact same shape as $x$ are generated and filled with masks where to flip the bits, in the following called \emph{fliptensors} (see figure \ref{fig:fliptensor}).
Here, $f_1$ is associated with the bit flips from 0 to 1, and $f_2$ vice versa.
Their elements are generated by sampling from a Bernoulli distribution with the given probabilities $p_1$ and $p_2$, respectively.
In the next step, bitwise operations are applied to the input bit tensor and the fliptensors simulating bit flip noise.
For the flip from 0 to 1, the operation should change the value of the input only if $x=0$ and $f_1=1$, else it should preserve it.
This can be realized by a bitwise OR operation, $x \vee f_1$.
On the other hand, the flip from 1 to 0 is represented by a conjunction, $x \wedge \neg f_2$.
This formula only changes the bit to 0 if $x=1$ and $f_2=1$.

In summary, the first operation is applied to an element $x_i$ if it equals 0, otherwise the second operation is used.
This can be re-formulated as
\begin{align*}
y_i & = \left(\neg x_i \wedge \left(x_i \vee f_{1,i}\right)\right) \vee \left(x_i \wedge \left(x_i \wedge \neg f_{2,i}\right)\right) \\
    & = \left(\neg x_i \wedge f_{1,i}\right) \vee \left(x_i \wedge \neg f_{2,i}\right) ,
\end{align*}
where $y_i$ is the output element with noise.
The number of bitwise operations can be reduced further leading to the equivalent expression
\begin{equation}
\label{eq:noise_operator}
y_i = \left(x_i \vee f_{1,i}\right) \veebar \left(x_i \wedge f_{2,i}\right) ,
\end{equation}
where $\veebar$ denotes the exclusive OR operation.
Using PyTorch, equation \ref{eq:noise_operator} generalizes to entire tensors, rendering explicit iterations over tensors unnecessary.

Equation \ref{eq:noise_operator} represents the operation performed by the PyTorch operator that we implemented to simulate bit flip noise.
If valid probabilities are provided to our program, the bits of the tensors of activations, weights and biases will be flipped according to these probabilities.
To increase flexibility, one can decide which of the three tensor types are subject to noise.

%% file: 05-experiments.tex
\section{Experiments on Noise Resilience}
\label{sec:experiments}

In this study we assume a processor with a simplified memory architecture.
All data is stored in resistive RAM, which is assumed to be large enough.
The processor writes directly to RRAM ignoring caching for now.
Unless stated otherwise, only one write pulse is performed, i.e., no write-check-correct scheme is applied by default.
Since write noise has been identified as the dominant noise factor, only activations are considered noisy.
Weights and biases are only read during inference and therefore, they are assumed to preserve their values.
As a concrete application, we investigate image classification using VGG neural networks \cite{Simonyan2015} of (convolutional) depths 11, 13, 16 and 19, and include batch normalization.
The dataset under consideration is CIFAR-10.

For each model architecture, an inference step over the entire validation dataset is performed.
Hereby, the write noise bit flip probabilities are swept over from $10^{-8}$ to $10^{-1}$.
This range is in line with the probabilities inferred from analyzing the distributions of cycle-to-cycle variabilities in Fantini et al.~\cite{Fantini2013}. 
The probabilities are assumed to be symmetric with respect to the direction of the flip.
Each run is repeated five times.

\begin{figure}
\begin{floatrow}
\ffigbox{%
  \includegraphics[width=0.45\textwidth]{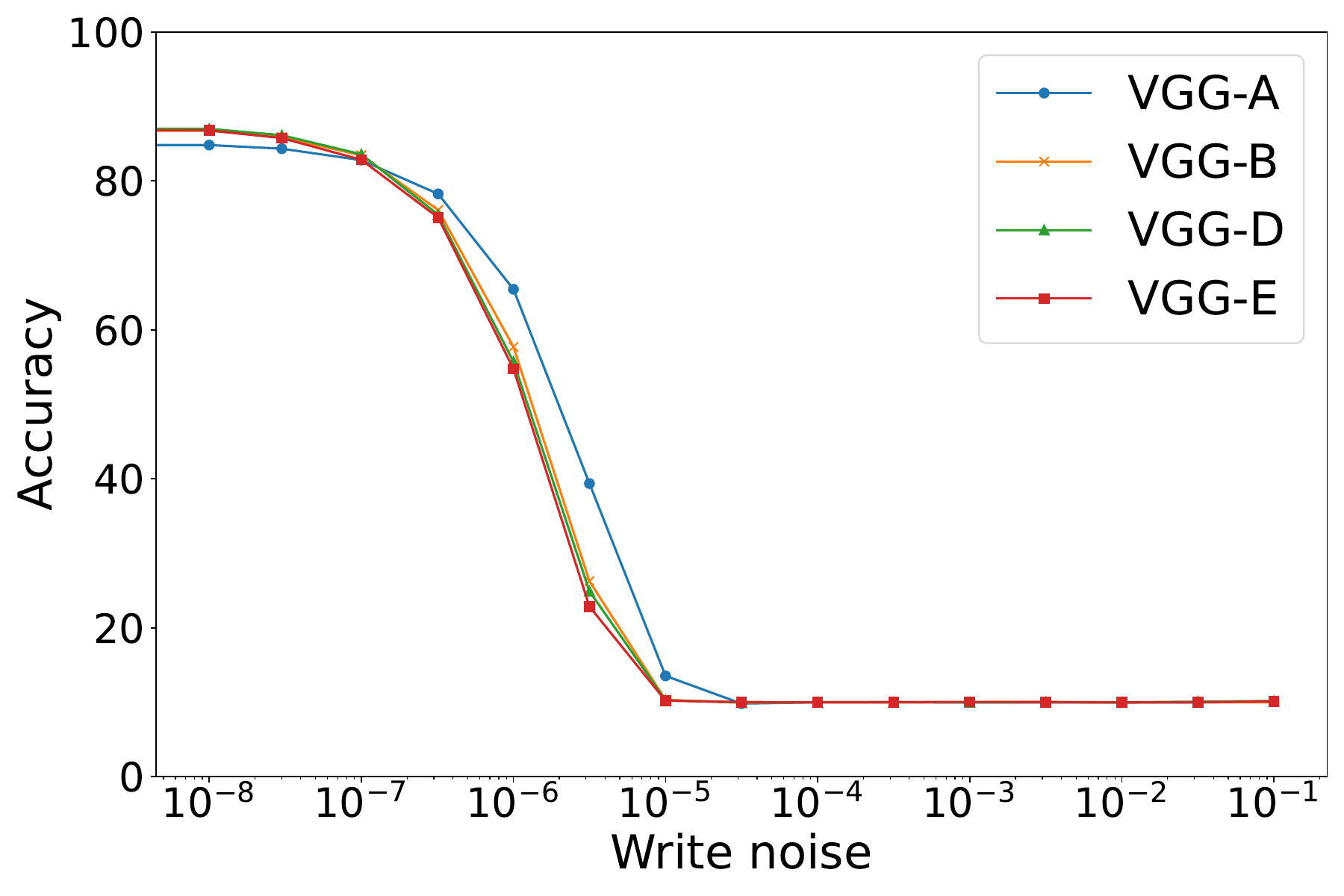}%
}{%
  \caption{Validation accuracy for VGGs of different depths for float32 data type.}%
  \label{fig:depths_cmp_float32}%
}
\capbtabbox{%
  \small
  \centering
  \begin{tabularx}{0.45\textwidth}{LRRR}
  VGG & float16 		 & float32 		 & float64  	 \\ 
  \midrule
  A    & \num{2.66} & \num{2.20} & \num{1.92} \\
  B    & \num{1.66} & \num{1.35} & \num{1.17} \\
  D    & \num{1.57} & \num{1.25} & \num{1.08} \\
  E    & \num{1.48} & \num{1.20} & \num{0.99} 
  \end{tabularx}
}{%
  \caption{Midpoint noise level $\mu$ (multiplied by $10^{-6}$) for different VGG depths and data types.}%
  \label{tab:midpoint_config}%
}
\end{floatrow}
\end{figure}

All four VGG networks show a similar behavior: starting from the trained baseline ($a_\mathrm{max}$), the accuracy drops to complete random guessing ($a_\mathrm{min}$) when increasing the probability of bit flips (see figure \ref{fig:depths_cmp_float32}).

The breakdown happens between probabilities of $10^{-7}$ and $10^{-5}$.
More precisely, the corresponding midpoint noise levels are summarized in table \ref{tab:midpoint_config}.
It is worth mentioning that the shallowest configuration (VGG-A with 11 convolutional layers) proves to be most robust, even though it has the lowest baseline accuracy.
This can be explained by the fact that this configuration has the smallest number of activations due to its depth, and hence, less noisy values in total.
The general trend of the accuracy, however, is the same.

\begin{figure*}
\centering
\includegraphics[width=1\textwidth]{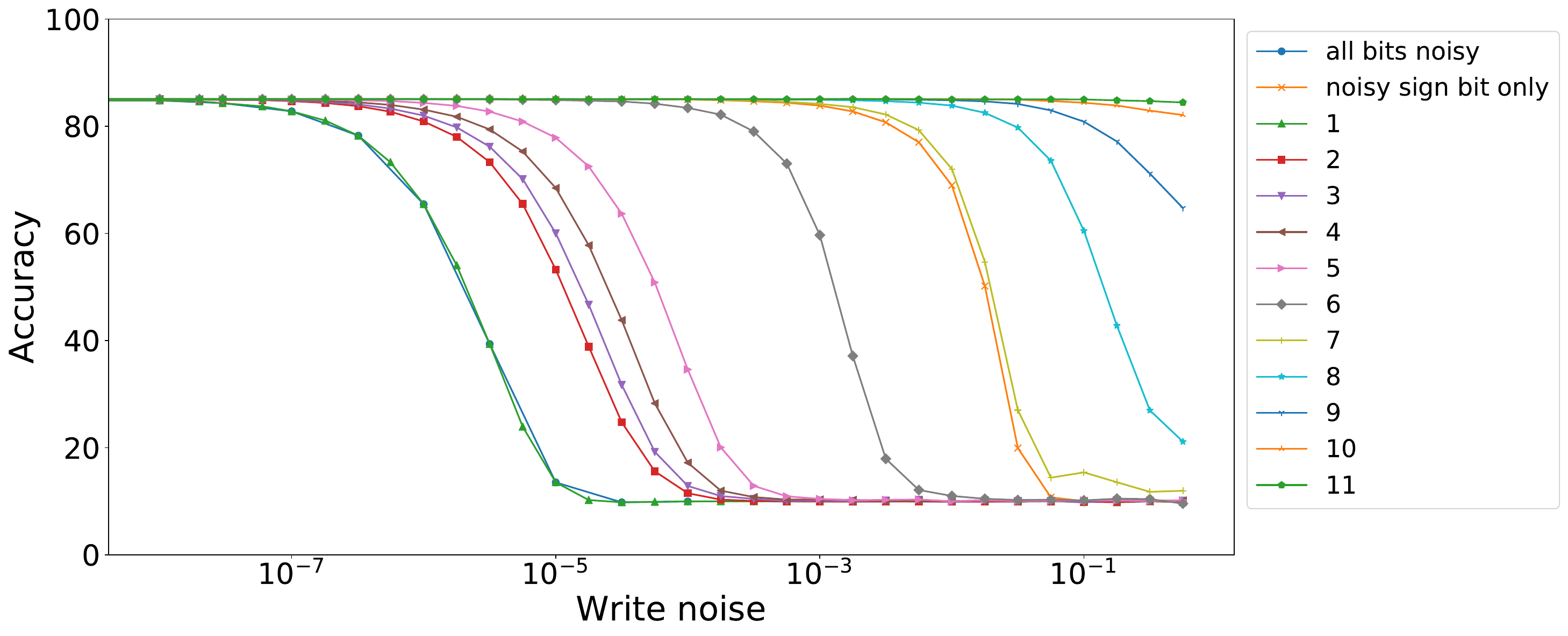}
\caption{Sweep over noisy bits for VGG-A on CIFAR-10. Explanation of the legend: the number is the position of the first noisy bit in big-endian format starting from 0. For instance, case "1" means that all but the sign bit are noisy. 
}
\label{fig:bitmask}
\end{figure*}

To examine the impact of noise more thoroughly, the sensitivity of the individual positions of the noisy bits is investigated, as it
is a reasonable hypothesis that bit flips of the higher significant bits have a stronger impact on the accuracy than bit flips of less significant bits.
Especially flipping the exponent bits of a floating-point number may change the value dramatically.
Hence, we investigate the accuracy drop for different bit masks in figure \ref{fig:bitmask}.
This is realized by reducing the number of noisy bits starting from the most significant bit, resulting in a unary code (a.k.a. thermometer code, for instance $3$ represented as $1110$).
Beginning with the most significant bit (the sign bit), bit flips are prohibited when sweeping over the probability range.
Remember that single-precision floating point (float32) is composed of one sign bit, $8$ exponent bits and $23$ mantissa bits.
In the first experiment, only the sign bit can be noisy, while in the second one every bit except the first two bits (sign bit and first exponent bit) can be noisy, and so on.
One clearly sees that the exponent bits are most influential on the accuracy.
Turning the noise completely off on the exponent and only allowing a noisy mantissa (case "9" in this figure) improves the robustness immensely: the accuracy drop then starts at a probability of about $10^{-1}$, with midpoint noise being off-chart.
Bit masks starting from bit 12 show no deviation from the baseline and are omitted\footnote{For additional results, please refer to subsection \ref{subsec:appendix_noisybits}.}.

%% file: 06-resilience.tex
\section{Improving Noise Resilience}
\label{sec:resilience}
As shown before, the inference accuracy drops to complete random guessing for a bit flip probability of about $10^{-5}$, i.e., when one out of \num{100000} bits is flipped.
Obviously, one could employ an enhancement strategy as write-check-correct cycles. 
This would require to read and compare the written value until the correct one is stored.
Another idea based on the previous results is to restrict noise to certain parts of a value, e.g., only the mantissa bits.
This could be realized by storing the exponent bits in SRAM or RRAM with better resilience.
Both methods, however, involve a change in the hardware set-up. 
On the other hand, in this work, we want to explore methods that are closer related to deep learning.

In the following, we examine the effect of reduction of the bitwidth of the floating-point data type, and also quantization.
While often considered in related work on resilience methods, to our surprise, noisy re-training did not yield any substantial improvement. 
We suspect that the large numerical variances as a result of noisy exponent bits result in too much noise than can be compensated by noisy re-training\footnote{Please refer to subsection \ref{subsec:appendix_retraining} in the appendix for details.}.

\subsection{Impact of Reduced-Precision Floating-Point Data Types}
\begin{figure}
\begin{floatrow}
\capbtabbox{%
  \centering
  \small

  \begin{tabularx}{0.35\textwidth}{LRR}
  & Exponent bits & Mantissa bits \\ 
  \midrule
  float16 	& 5 	& 10 \\
  bfloat16 	& 8 	& 7  \\
  float32 	& 8 	& 23 \\
  float64 	& 11 	& 52
  \end{tabularx}

}{%
  \caption{Comparison of different floating-point types.}%
  \label{tab:floats}%
}
\ffigbox{%
  \includegraphics[width=0.5\textwidth]{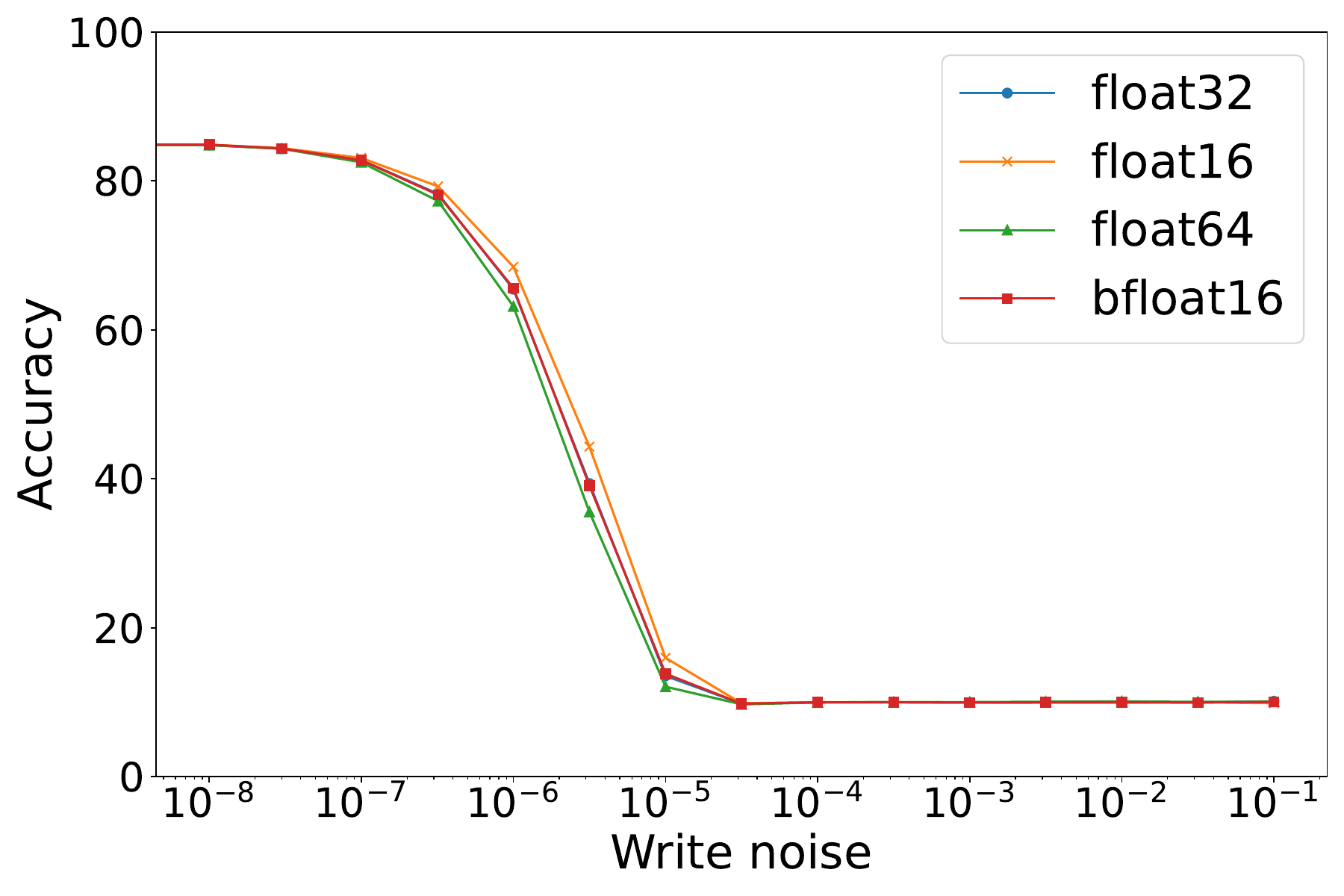}
}{%
  \caption{Validation accuracy for different floating-point data types (for inference)}%
  \label{fig:vgg_A_dtypes}%
}
\end{floatrow}
\end{figure}

To emphasize the importance of the underlying bit representation, the write noise sweep is carried out for floating-point types of four different bitwidths: float16, bfloat16, float32 and float64.
The important difference is the varying number of exponent bits (see table \ref{tab:floats}).
Results from figure \ref{fig:vgg_A_dtypes} suggest that the noise level at which the accuracy drop occurs depends on the number of exponent bits.
By that reason, using float16 leads to a slightly better robustness, whereas using double precision decreases it even further.
With respect to robustness, bfloat16 and float32 are interchangeable: as both of them use an exponent width of 8 bits, the inference accuracy coincides for all noise probabilities.

\subsection{Integer Quantization}
\label{subsec:quantization}
The next resilience method we investigate is integer quantization.
The idea is to map the weights, biases and activations from floating-point numbers to an integer range of lower bit width (here: eight bits).
By this procedure, not only the size of the model can be reduced, but also the number of bits with high impact on the accuracy.
We employed static post-training quantization on all activations, weights and biases.
For simplicity reasons, quantization-aware training or re-training was not considered here.

\begin{figure}
\begin{floatrow}
\ffigbox{
  \includegraphics[width=0.45\textwidth]{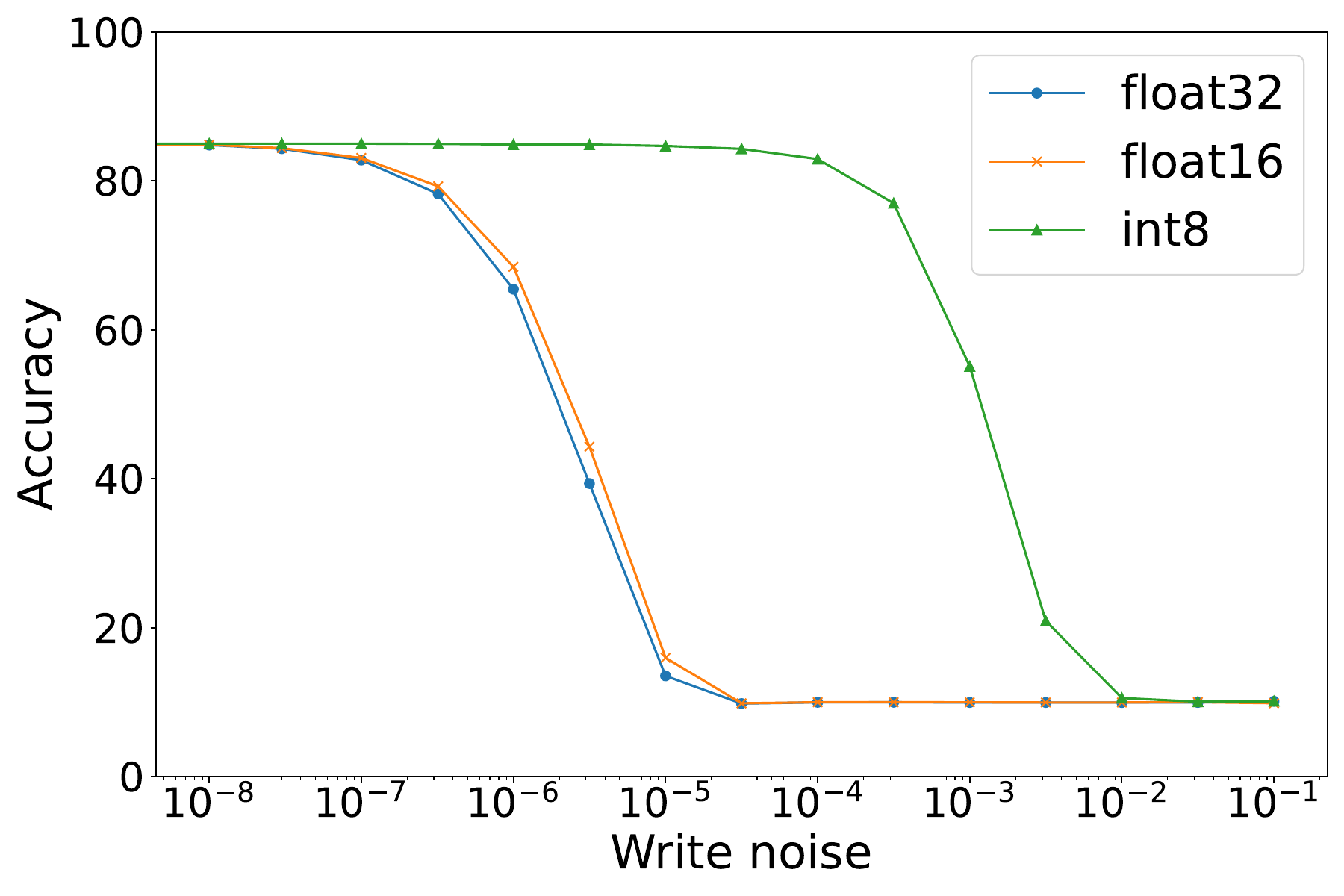}
}{
  \caption{Comparison of VGG-A/CIFAR-10 accuracy for 32- and 16-bit floating-point, was well as 8-bit integer quantization.}
  \label{fig:vgg_A_quant}
}
\capbtabbox{
  \centering
  \small
  \begin{tabularx}{0.45\textwidth}{XRR}
   Network type & Midpoint noise level $\mu$ & Improvement over float32 \\ \midrule
   A            & \num{1.23e-3}              & 557x \\
   B            & \num{1.58e-3}              & 1167x \\
   D            & \num{1.45e-3}              & 1159x \\ 
   E            & \num{1.27e-3}              & 1054x
  \end{tabularx}
}{
  \caption{Midpoint noise for quantized networks.}
  \label{tab:midpoint_noise_quant}
}
\end{floatrow}
\end{figure}

In figure \ref{fig:vgg_A_quant} the accuracy results for VGG-A/CIFAR-10 for float32 and float16 data types are compared to 8-bit integer quantization.
Strikingly, quantization shifts the noise probability at which the accuracy drop happens by three orders of magnitude.
This effect can be generalized to the other depths of VGG, as shown in table \ref{tab:midpoint_noise_quant}.
In summary, this supports and highlights the importance of avoiding noise on exponent bits.

%% file: 07-summary.tex
\section{Summary}

The key question that motivated the present work is how much noise in memory operations can be tolerated by plain ML architectures, and how this can be improved by resilience methods. 
While limiting this early study to convolutional neural networks based on the VGG architecture and CIFAR-10 as an image classification task, we came to the following insights:

Noisy memory accesses for floating-point-based model architectures are disastrous, however, limiting the amount of noisy exponent bits is a very promising countermeasure.\footnote{Please refer to subsection~\ref{subsec:appendix_midpoint_noise} in the appendix for a detailed study of midpoint noise.}
While to our surprise noisy re-training did not result in any notable improvements on the resilience against noise, quantizing to an integer-based number format was key in improving resilience. 

In this regard, consider that $99\%$ of peak accuracy might be tolerable from an application point of view ("$99\%$-noise").
For VGG-A based on float32, this would allow to tolerate a noise level of \num{1.15E-07}, while post-training integer quantization would improve this robustness to \num{1.02E-04}.
Considering midpoint noise levels for this example, $\mu$ would improve from \num{2.20E-06} to \num{1.23E-03}. 
Generally, an improvement of about three orders of magnitude holds true for all considered VGG models, with VGG-A showing the smallest improvement (midpoint noise improving by a factor of \num{5.57E+02}) and VGG-B as well as VGG-D the largest ($99\%$-noise improving by a factor of \num{3.25E+03}).
Thus, by choosing a good compromise among quantization and model accuracy, the amount of tolerable noise can be substantially increased.
Furthermore, we observe strong scaling effects in the midpoint noise level: the larger the total number of noisy exponent bits in the model architecture, the lower this metric is, and, thus, the less noise resilience there is in a given model architecture.

Future work will extend this work to understand the benefits of tolerating noise towards costs in terms of time, area and energy.
Of similar interest is trading among performance, power and reliability of resistive memory units, 
including a quantification of those metrics.

%% file: 08-appendix.tex
\section{Appendix}

\subsection{Midpoint noise level for noisy exponent bits (section \ref{sec:experiments})}
\label{subsec:appendix_noisybits}

The findings from figure \ref{fig:bitmask} (sweep over noisy bits for VGG-A on CIFAR-10) are also reflected in the corresponding midpoint noise levels in table \ref{tab:midpoint_exponent}, as there is a strong correlation between the significance of the noisy bits and the midpoint noise.
For the mantissa bits, the fit to equation \eqref{eq:logistic} fails as the accuracy drop occurs only for bit flip probabilities that are higher than the probed ones (larger than \SI{50}{\percent}).
Hence, no midpoint noise levels can be reported.

\begin{table}
\centering
\small
\caption{Midpoint noise level $\mu$ for different noisy MSBs of the exponent.}
\label{tab:midpoint_exponent}
\begin{tabularx}{0.85\textwidth}{LRRRR}
Exponent bit  & VGG-A & VGG-B & VGG-D & VGG-E \\ \midrule
1 & \num{2.21e-06} & \num{1.37e-06} & \num{1.26e-06} & \num{1.20e-06} \\
2 & \num{1.21e-05} & \num{6.50e-06} & \num{6.09e-06} & \num{5.70e-06} \\
3 & \num{1.64e-05} & \num{9.00e-06} & \num{8.48e-06} & \num{7.98e-06} \\
4 & \num{2.57e-05} & \num{1.46e-05} & \num{1.39e-05} & \num{1.34e-05} \\
5 & \num{5.93e-05} & \num{3.76e-05} & \num{4.05e-05} & \num{4.18e-05} \\
6 & \num{1.33e-03} & \num{1.24e-03} & \num{1.40e-03} & \num{1.30e-03} \\
7 & \num{2.00e-02} & \num{1.74e-02} & \num{2.11e-02} & \num{1.95e-02} \\
8 & \num{1.56e-01} & \num{1.13e-01} & \num{1.48e-01} & \num{1.29e-01}
\end{tabularx}
\end{table}

\subsection{Impact of Noisy Re-Training (section \ref{sec:resilience})}
\label{subsec:appendix_retraining}
Performing further training steps of the model including noise injection to a certain extent proved itself to be an effective countermeasure for Gaussian noise on the weights \cite{Zhou2020}.
Given that knowledge, noisy re-training was carried out for VGG-A (the most shallow and most resilient configuration).
For ten training epochs, the model was re-trained for six different noise probabilities ranging from $10^{-6}$ to $10^{-1}$.
The noise was only applied in the forward path.
As re-training fails when allowing noise on all bits, we applied different bitmasks to the exponent starting from the third exponent bit until the entire exponent is noiseless and only the mantissa bits are noisy.

However, noisy re-training does not turn out to be effective.
Requiring a baseline accuracy, i.e., without noise, of more than \SI{80}{\percent}, the midpoint noise increases only slightly to \num{2.35e-06}.
The corresponding bitmask started at the sixth exponent bit and the bit flip probability for re-training was \num{1e-3}.
Even by loosening the restriction on the accuracy to be larger than \SI{50}{\percent} yields a midpoint noise of only \num{2.52e-06} at a maximum accuracy of \SI{51.05}{\percent}.
As a comparison, without noisy re-training VGG-A achieved a midpoint noise level of \num{2.20e-6} and a noiseless accuracy of \SI{85.07}{\percent}.

\subsection{Correlation of Bit Flip Probability, Data Type, and Total Number of Noisy Exponent Bits (section \ref{subsec:quantization})}
\label{subsec:appendix_midpoint_noise}

\begin{figure}
\centering
\includegraphics[width=0.75\textwidth]{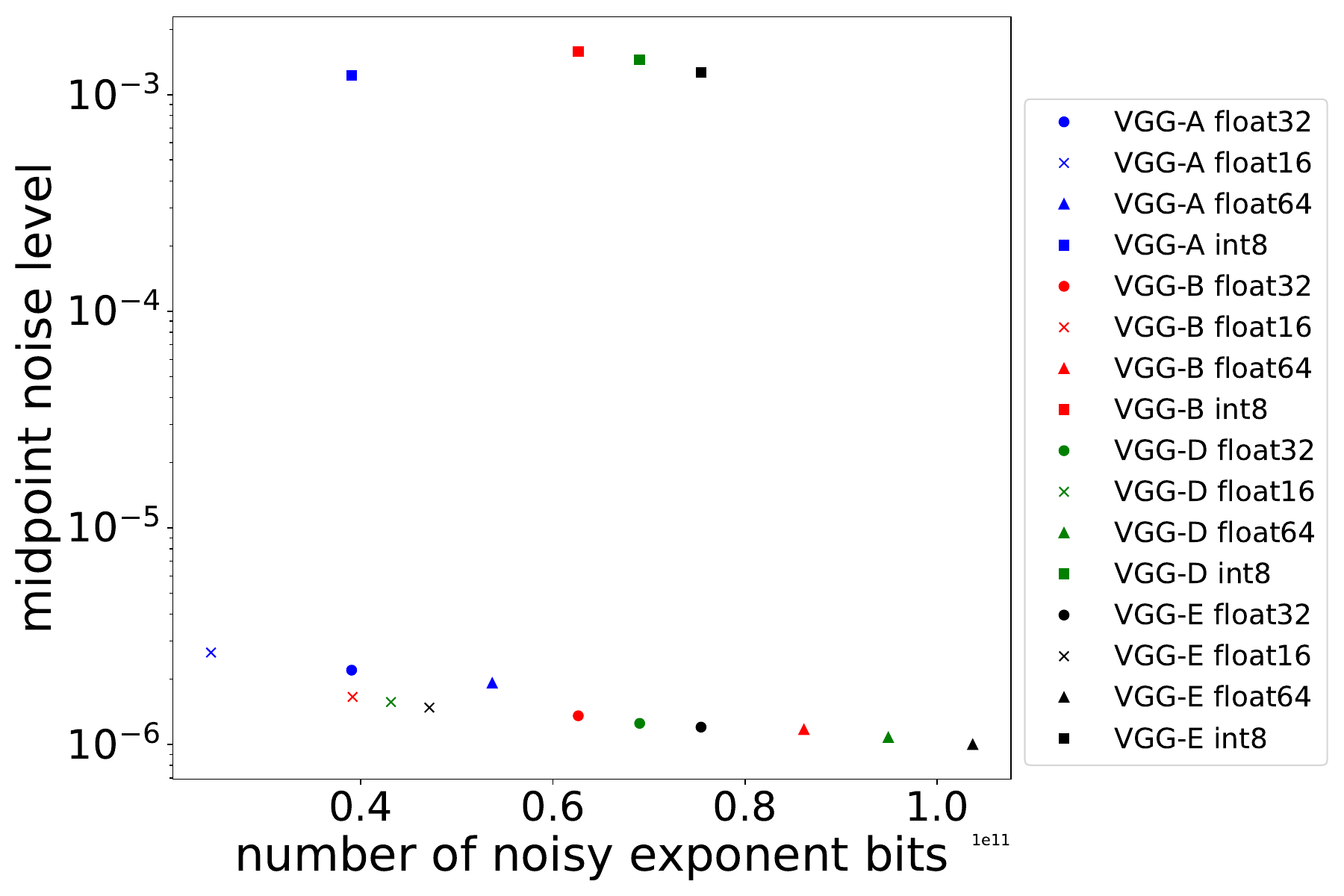}
\caption{Midpoint noise levels as a function of number noisy bits for different data types.}
\label{fig:p_at_half_max_bits}
\end{figure}

In figure \ref{fig:p_at_half_max_bits} we report the bit flip probability $p$ required to achieve half-maximum accuracy for the different VGG variants, over the total number of noisy exponent bits in the model architecture, thereby allowing a comparison between the VGG variants.
At the bottom of the figure are the values of the midpoint noise for the networks using floating-point numbers for the activations, whereas the data points at the top refer to the same networks with quantized activations.

This result can be explained by the mapping procedure of quantization.
Given a floating-point number $r$, the quantized version of this number is calculated as $Q\!\left(r\right) = \mathrm{round}\!\left(\frac{r}{S} + Z\right)$.
Here, $S$ is the scaling factor, defined as 

\begin{equation}
S = \frac{\beta - \alpha}{\beta_q - \alpha_q} \, ,
\end{equation}
where $\left[\alpha, \beta\right]$ is the range of the inputs and $\left[\alpha_q, \beta_q\right]$ is the range of the quantized values.
$Z$ is the zero point:
\begin{equation}
Z = -\left(\frac{\alpha}{S} - \alpha_q\right) \, .
\end{equation}

The quantized values are confined to a much tighter range $\left[\alpha_q, \beta_q\right]$ than the inputs.
Also, as bit flips in integers only add or subtract a power of 2, conversely to floating-point numbers, for which flips of exponent bits change the entire scale of the number, their impact on the accuracy is drastically reduced.

Finally, we observe scaling effects in the midpoint noise probability (figure \ref{fig:p_at_half_max_bits}):
the larger the number of noisy bits, the lower is this probability.
This implies that the accuracy drop happens for smaller probabilities, meaning a lower robustness.
Hence, the shallowest VGG configuration is favorable, which is inline with the observations in section \ref{sec:experiments} (cf.\ table \ref{tab:midpoint_config}).
Floating-point numbers and quantized numbers define different scales (top: quantized; bottom: floating-point), but the general trend is visible for all data types.